\documentclass[preprint,3p,times]{elsarticle}

\usepackage{graphicx}
\usepackage{booktabs}
\usepackage{amsmath}
\usepackage{siunitx}
\usepackage{xspace}
\usepackage[hidelinks]{hyperref}
\usepackage{orcidlink}
\newcommand{\datasetName}{Precision Agriculture Farm Machinery Autonomous
Driving\xspace}
\newcommand{\nModalityArms}{four\xspace}    
\newcommand{\nSuperSites}{three\xspace}     
\newcommand{\nModelConfigs}{three\xspace}   
\newcommand{\nSeeds}{five\xspace}           
\newcommand{\nTargetClasses}{five\xspace}   
\newcommand{\nGroundTruths}{three\xspace}   
\newcommand{\nCorpusEnvironments}{three\xspace} 
\newcommand{\nEnvironments}{two\xspace}     
\newcommand{\nControls}{two\xspace}         
\newcommand{\controlFloor}{0.30\xspace}     
\newcommand{\alphaLevel}{0.05\xspace}       
\newcommand{\restrictedEnv}{paddy\xspace}   
\newcommand{\backboneSmall}{MiT-B0\xspace}  
\newcommand{\backboneLarge}{MiT-B2\xspace}  
\newcommand{\archSecond}{UPerNet/\allowbreak ConvNeXt-V2-T\xspace}  

\newcommand{\ranusVideos}{50\xspace}      

\newcommand{\mergeRadiusM}{11}              
\newcommand{\trainEpochs}{30\xspace}        
\newcommand{\trainBatch}{16\xspace}         
\newcommand{\cropSize}{512\xspace}          
\newcommand{\inputWidth}{960}               
\newcommand{\inputHeight}{544}              
\newcommand{\learningRate}{3\times10^{-4}}  
\newcommand{\weightDecay}{10^{-4}}          
\newcommand{\frameStride}{5\xspace}         
\newcommand{\samplerCap}{8\xspace}          
\newcommand{\fusionParamFactor}{1.8\xspace} 
\newcommand{\bootResamples}{\num{1e6}}      
\newcommand{\boundarySEs}{3\xspace}         
\newcommand{\ciEndpointFloor}{\num{5e-4}}   

\newcommand{\nPairedFrames}{79029}  
\newcommand{\nOfficialOverlaps}{270}  
\newcommand{\nIndexFrames}{15790}  
\newcommand{\blockedValSharePct}{27.9}  

\newcommand{\nirEdgeOfficial}{0.339\xspace}  
\newcommand{\nirEdgeBlocked}{0.015\xspace}  
\newcommand{\nirEdgeIsecFrames}{322}  
\newcommand{\rgbEdgeOfficial}{0.009\xspace}  
\newcommand{\nirRgbEdgeOfficial}{\ensuremath{+0.127}}  
\newcommand{\nirRgbEdgeBlocked}{\ensuremath{-0.013}}  
\newcommand{\nirRgbEdgeBlockedCI}{\ensuremath{[-0.028,\,-0.000]}}  

\newcommand{\nirLumBeforeBlocked}{\ensuremath{+0.020}}  
\newcommand{\nirLumBeforeBlockedCI}{\ensuremath{[+0.000,\,+0.048]}}  
\newcommand{\nirLumBeforeSiteP}{0.92}  
\newcommand{\nirLumBeforeSiteLow}{\ensuremath{-0.038}}  
\newcommand{\nirLumAfterBlocked}{\ensuremath{+0.030}}  

\newcommand{\fusionColourBlocked}{\ensuremath{+0.033}}  
\newcommand{\fusionColourBlockedCI}{\ensuremath{[+0.017,\,+0.048]}}  
\newcommand{\fusionColourFurrow}{\ensuremath{+0.076}}  
\newcommand{\fusionColourFurrowCI}{\ensuremath{[+0.014,\,+0.107]}}  
\newcommand{\fusionColourSiteTwo}{\ensuremath{+0.032}}  
\newcommand{\fusionColourSiteOne}{\ensuremath{+0.010}}  
\newcommand{\fusionColourSiteZero}{\ensuremath{-0.075}}  
\newcommand{\fusionColourSiteMean}{\ensuremath{-0.011}}  
\newcommand{\fusionColourSiteP}{0.77}  
\newcommand{\siteLevelTCrit}{4.30}  

\newcommand{\nirSeenIsec}{0.172\xspace}  
\newcommand{\nirUnseenIsec}{0.016\xspace}  
\newcommand{\nirSeenUnseenIsec}{\ensuremath{+0.067}}  
\newcommand{\rgbSeenUnseenIsec}{\ensuremath{-0.008}}  
\newcommand{\lumSeenUnseenIsec}{\ensuremath{+0.027}}  
\newcommand{\nirSeenUnseenOwn}{\ensuremath{+0.119}}  
\newcommand{\rgbSeenUnseenOwn}{\ensuremath{+0.059}}  
\newcommand{\lumSeenUnseenOwn}{\ensuremath{+0.139}}  
\newcommand{\suNirRgb}{\ensuremath{+0.076}}  
\newcommand{\suNirRgbCI}{\ensuremath{[+0.017,\,+0.142]}}  
\newcommand{\suNirLum}{\ensuremath{+0.040}}  
\newcommand{\suNirLumCI}{\ensuremath{[+0.007,\,+0.093]}}  
\newcommand{\suLumRgb}{\ensuremath{+0.035}}  
\newcommand{\suLumRgbCI}{\ensuremath{[+0.002,\,+0.075]}}  
\newcommand{\suLumRgbOwn}{\ensuremath{+0.080}}  
\newcommand{\suLumRgbOwnCI}{\ensuremath{[-0.013,\,+0.195]}}  
\newcommand{\suSeqIsec}{4}  
\newcommand{\suSeqIsecMeans}{35}  
\newcommand{\suSeqOwn}{6}  

\newcommand{\camShiftPx}{7}  
\newcommand{\camShiftControlPx}{1}  
\newcommand{\annotSiteIdDelta}{\ensuremath{+0.006}}  
\newcommand{\annotSiteIdDeltaCI}{\ensuremath{[-0.009,\,+0.021]}}  

\newcommand{\eThreeControlSiteZero}{0.248\xspace}  
\newcommand{\eThreeControlSiteOne}{0.806\xspace}  
\newcommand{\eThreeControlSiteTwo}{0.801\xspace}  
\newcommand{\fieldShareSiteZeroPct}{66.4}  
\newcommand{\sepCommittedM}{761}  
\newcommand{\sepPastM}{0.5}  
\newcommand{\eTwoPooled}{\ensuremath{+0.064}}  
\newcommand{\eTwoPooledCI}{\ensuremath{[+0.023,\,+0.104]}}  
\newcommand{\eTwoSiteLevelP}{0.27}  
\newcommand{\eOneRgbSiteZeroCI}{\ensuremath{[-0.002,\,+0.002]}}  

\newcommand{\gOneMinFoldMean}{0.392\xspace}  
\newcommand{\gTwoNirEdgeSiteZero}{\ensuremath{+0.245}}  
\newcommand{\gThreePooled}{\ensuremath{+0.080}}  
\newcommand{\gThreePooledCI}{\ensuremath{[+0.028,\,+0.137]}}  
\newcommand{\gThreePooledN}{7317}  
\newcommand{\gThreeSiteZero}{\ensuremath{+0.140}}  
\newcommand{\gThreeSiteOne}{\ensuremath{-0.027}}  
\newcommand{\gThreeSiteTwo}{\ensuremath{+0.073}}  
\newcommand{\gThreeSiteLevelP}{0.33}  

\journal{Computers and Electronics in Agriculture}

\begin{document}

\begin{frontmatter}

\title{Forward-Facing Near-Infrared Adds Little to Colour for Farm-Machinery
Traversability: A Site-Disjoint Evaluation of Sensor-Dependent Spatial
Leakage}

\author[a]{Sungwoo Kang\texorpdfstring{\,\orcidlink{0009-0004-0037-7593}}{}}
\ead{krml919@korea.ac.kr}
\affiliation[a]{organization={Department of Electrical and Computer
Engineering, Korea University}, city={Seoul}, country={Republic of Korea}}

\begin{abstract}
Forward-facing near-infrared imaging is proposed to complement colour cameras
for agricultural traversability, yet whether the extra band supplies
information colour does not already carry is unresolved, because the
benchmarks behind it split spatially autocorrelated imagery without holding
locations apart. We evaluate \nModalityArms{} input
configurations, colour, near-infrared, a capacity-matched luminance control,
and their fusion, on the daylight imagery of the AI Hub \datasetName{} corpus
under protocols that hold the recording sites apart. No configuration then
beats colour consistently on any of the \nTargetClasses{} traversability
target classes, and the two effects favouring the second camera at a single
held-out site both fail to replicate when the held-out site is rotated. For a
forward-facing sensor suite on daytime paddy and dry-field driving, the second
camera is unproven. Under the corpus's own sequence-level split, near-infrared
instead appears to succeed on a paddy-boundary class that colour segments
poorly. That advantage is site leakage, and near-infrared is the configuration
it inflates most on that class. Correcting the split removes near-infrared's
lead rather than lowering every configuration alike. A location-leaking
evaluation can reorder the sensors a procurement decision depends on, and a
single held-out site can report a site-specific effect as general.
\end{abstract}


\end{frontmatter}

\section{Introduction}\label{sec:intro}

Autonomous farm machinery must decide where it can drive safely. The decision
reduces to segmenting the scene ahead into traversable and non-traversable
ground. Colour cameras supply the imagery for that segmentation. Near-infrared
imaging has been proposed as a complement. Chlorophyll-bearing leaves reflect
strongly beyond the visible range. The reflectance separates vegetation from
soil more sharply than the visible channels do
\citep{chebrolu2017agricultural}. Soil reflectance also darkens with moisture
across the shortwave range \citep{lobell2002moisture}. Segmentation studies on
road and outdoor scenes report accuracy gains from the band
\citep{choe2018ranus,valada2016deep}. Whether the extra band supplies
traversability signal that colour does not already carry is a procurement
question as much as a scientific one. A near-infrared camera adds cost and
calibration effort, and only a non-redundant gain repays them.

Neither benchmark behind those reported gains separates training from
evaluation by location. The urban corpus draws its annotated key frames from
\ranusVideos{} driving videos and splits the frames by scene category, stating
rather than showing that its training and test images are uncorrelated
\citep{choe2018ranus}. The forested benchmark describes no split at all
\citep{valada2016deep}. Frames captured along a driving trajectory are
spatially autocorrelated \citep{roberts2017cv,ploton2020spatial}. A partition
that ignores location places neighbouring and near-duplicate views on both
sides of it. Spatially unaware validation inflates the measured accuracy of
convolutional segmentation networks \citep{kattenborn2022spatially}. It also
promotes geolocation to the most important predictor in ecological models
\citep{meyer2019importance}. A sensor advantage reported under such a split
measures how well a model memorises sites rather than how well it transfers to
an unseen field.

A deployed machine drives on fields absent from any training set. A benchmark
that rewards site memorisation can rank sensors in an order that reverses on
such a field. We study the question on the AI Hub \datasetName{} corpus, which
pairs the two bands over rice-paddy and dry-field driving and is partitioned
into training and validation over whole recording sequences. Separate
sequences revisit the same plots. The sequence-level split leaves one
location's frames on both sides of it. Under that split near-infrared appears
to succeed on a paddy-boundary class that colour segments poorly. The apparent
gain cannot be distinguished from the location overlap without protocols that
hold the recording sites apart.

We evaluate \nModalityArms{} input configurations under the two protocols. The
configurations are colour, near-infrared, a capacity-matched luminance control
that is the greyscale of the colour image, and the fusion of colour and
near-infrared. The luminance control separates a spectral-band gain from the
extra intensity map that near-infrared also supplies. A near-infrared gain
over the control is an effect of the band and not of added brightness. The
protocols train and evaluate on disjoint super-sites, the location clusters we
recover from the recorded GPS tracks. One holds a block of sites out, and the
other rotates which site is held out. Both protocols are set against the
sequence-level split. The accuracy inflation attributable to leakage is
measured directly. Each configuration is scored against \nGroundTruths{}
ground truths, the colour annotations, the near-infrared annotations, and
their pixel intersection, the last of which is free of annotation-side
leakage.

This paper makes two contributions. First, near-infrared beats colour on none
of the \nTargetClasses{} traversability target classes under evaluation that
holds the recording sites apart. Fusing the two recovers no consistent gain
either. Two comparisons in the second camera's favour do satisfy their
criterion at the single site the site-blocked protocol holds out, fusion over
colour and near-infrared over the luminance control, and both change sign once
the held-out site is rotated. For daytime paddy and dry-field driving, the
second camera is unproven for this task. Second, the apparent gain is site
leakage, and the leakage is sensor-dependent. The near-infrared advantage on
the paddy edge is recognition of the site. The advantage tracks whether a
model saw the evaluation site rather than how hard the site is to segment. It
does not survive holding the sites apart. Leakage of that kind is a property
of how a corpus is partitioned rather than of how it was recorded, and any
partition drawn without a location constraint produces it on spatially
autocorrelated data. The size of the effect is specific to the sensor.
Near-infrared gains more than colour from having seen an evaluation site
during training, and more than a capacity-matched luminance control, at both
held-out sites on which the class is evaluable and on \nModelConfigs{} model
configurations spanning two architecture families. Correcting the evaluation
removes near-infrared's lead rather than lowering every configuration alike.
Leakage distorts which predictors a model appears to use
\citep{meyer2019importance}. Here it distorts which sensor is bought or
omitted.

The remainder of the paper is organised as follows. Section~\ref{sec:related}
places the study among work on spatial leakage, sensor redundancy, and
traversability datasets. Section~\ref{sec:methods} describes the corpus, the
input configurations, the protocols, and the statistical tests.
Section~\ref{sec:results} reports the results, and
Sections~\ref{sec:discussion} and~\ref{sec:conclusion} discuss their scope and
implications.

\section{Related work}\label{sec:related}

Spatial autocorrelation makes independent random splits inappropriate for
geospatial prediction, and the resulting optimistic bias is well quantified.
Neighbouring samples share values through spatial proximity. A random test
point frequently has a near-duplicate in the training set. Standard
cross-validation then reports an accuracy the model cannot reproduce away from
the sampled locations \citep{roberts2017cv,ploton2020spatial}. Blocked and
leave-one-location-out cross-validation are the established remedies
\citep{roberts2017cv,karasiak2022spatial}, and the bias they correct extends
to modern architectures. \citet{kattenborn2022spatially} report substantially
inflated accuracy for a convolutional segmentation network validated on tiles
from the image acquisitions it trained on. The same failure is documented
outside geospatial work as leakage, a recurring cause of overstated results
answered by a diagnostic audit paired with a corrected protocol
\citep{kaufman2012leakage,kapoor2023leakage}. We bring that audit structure to
forward-facing agricultural traversability with paired colour and
near-infrared imagery, a setting in which it has not been applied, and treat
one public corpus as a case study of an effect that any split drawn without a
location constraint produces on spatially autocorrelated data.

Leakage costs more than the accuracy level. \citet{meyer2019importance} show
that it distorts which predictors a model appears to use. Geolocation
dominates the variable importance their random forest reports, and a spatially
blocked selection drops the same variables as counterproductive. This study
extends that distortion from predictors within one feature set to whole sensor
modalities, and to forward-facing traversability rather than map-scale
prediction. \citet{wadoux2021spatial} find blocked validation grossly
pessimistic as an estimate of map accuracy over a sampled region, because it
scores the model only where calibration data are absent. We measure transfer
to an unseen site, not map accuracy. The regime they identify as pessimistic
is the regime a deployed machine meets.

\citet{valada2016deep} and \citet{choe2018ranus} report segmentation gains
from a near-infrared band on road and outdoor scenes. The reported
comparisons leave open whether
a near-infrared gain is non-redundant with a same-capacity brightness channel,
because a model given near-infrared reads one intensity map whereas a
colour-only model reads three chromatic ones. We remove that confound using a
capacity-matched luminance control, the greyscale of the colour image. The
comparison of near-infrared against the control holds the input structure
fixed and varies the band.

The off-road corpus GOOSE pairs forward-facing colour with near-infrared
\citep{mortimer2024goose}. Its fixed split is balanced across seasons, weather,
and semantic classes rather than across locations. In agriculture, aerial and
downward-facing cameras record the two bands together
\citep{chebrolu2017agricultural,chiu2020agriculture}. Neither viewpoint shows
the ground ahead of a moving machine. The AI Hub \datasetName{} corpus pairs
forward-facing colour with near-infrared over rice-paddy and dry-field driving.
We build a leakage-aware split on it.

\section{Material and methods}\label{sec:methods}

\subsection{Dataset, annotations, and site recovery}\label{sec:materials}

The AI Hub \datasetName{} corpus provides forward-facing paired colour and
near-infrared imagery recorded from agricultural machinery driving through
rice paddies, dry fields, and orchards in the Republic of Korea. A colour
camera and a co-mounted near-infrared camera view the same scene ahead of the
vehicle. Each modality carries its own pixel-level semantic annotation, and
the two annotation sets do not cover an identical list of frames. This study
works from the frames annotated on both. The dataset also provides LiDAR
returns, which this comparison of colour with near-infrared does not use.

The dataset documentation gives the design intent of the second camera. Pairing
the two bands is intended to make object and obstacle estimation usable in
daytime and at dusk, and the documentation names detection of the
cultivated-plot boundary as one of the two tasks the pairing is meant to serve.
The corpus includes the class the pairing was designed for, the paddy edge,
among the \nTargetClasses{} target classes evaluated here. The same
documentation scopes the intent to daylight and dusk. Every paired colour and
near-infrared frame in the corpus is a daylight recording under clear weather,
and the release contains no night imagery to evaluate.

The annotation schema targets the ground classes an agricultural robot must
distinguish to navigate. Among the ground classes, \nTargetClasses{} carry the
traversability signal, namely the untilled paddy surface, the tilled paddy
surface, the paddy edge, the field furrow, and the field levee. The first two
are the same ground before
and after tillage. The annotation guideline distinguishes them by the colour of
the worked plot, and the released English class keys name them
\texttt{paddy\_before\_driving} and \texttt{paddy\_after\_driving}.
We reserve \nControls{} further classes as controls on which a spectral-band
advantage is not predicted. The controls are the farm access road leading to a
working plot and trees standing within the plot. The standing-water and
dry-field-crop classes had no training instances in the scene-matched frames
and are excluded. A standing rice-crop class appears in the taxonomy but is
neither a traversability target nor a control, and we report it only for
completeness. All remaining pixels form the background.

The recordings divide into \nCorpusEnvironments{} driving environments, rice
paddy, dry field, and orchard. The orchard recordings carry none of the
\nTargetClasses{} traversability classes, whose taxonomy is defined over paddy
and field ground. This study uses the \nEnvironments{} field-crop environments
and drops the orchard material. The remaining recordings also divide spatially
into super-sites, the location clusters recovered from the recorded GPS
tracks. A single super-site can contribute frames to both environments, and
the environments are not spread evenly across the super-sites. The composition
governs which held-out-site evaluations are feasible on this corpus.

We recover the super-sites by single-linkage clustering of the sequence GPS
tracks. Two sequences join the same super-site when the exact point-to-point
distance between their GPS tracks, computed using a KD-tree, is within
\SI{\mergeRadiusM}{\metre}. Clustering the corpus this way returns
\nSuperSites{} super-sites.

The corpus is distributed with a training and validation partition drawn over
whole recording sequences, and no sequence appears on both sides of it. We
call it the sequence-level split. Separate sequences nonetheless revisit the
same plots. The split leaves one location's frames on both sides of the
partition. A partition drawn without a spatial constraint does not control
that dependence \citep{roberts2017cv,ploton2020spatial}. We take the
sequence-level split as one evaluation condition and set it against the
site-disjoint protocols. The accuracy attributable to shared locations is then
measured directly.

\subsection{Input configurations and models}
We compare \nModalityArms{} input configurations that differ in the image
bands presented to the network. The colour configuration receives the three
visible channels. The near-infrared configuration receives the single
near-infrared band, replicated across three channels so that the
ImageNet-initialised stem is reused unchanged. The fusion configuration
receives colour and near-infrared as two three-channel streams. The fourth
configuration is a luminance control, the Rec.~601 luma of the colour image,
which is a fixed weighted sum of the red, green, and blue channels, replicated
across three channels in the same way.

The near-infrared configuration presents one intensity map whereas the colour
configuration presents three chromatic ones. A near-infrared advantage could
then reflect that difference in input structure rather than the spectral band.
The luminance control has the same single-map structure and is a deterministic
function of the colour image. It carries no near-infrared information.
Networks exploit whatever incidental cue distinguishes two inputs
\citep{geirhos2019imagenet,geirhos2020shortcut}. The input structure must be
held fixed before a band effect can be read. A near-infrared gain over the
control is an effect of the spectral band rather than of the single-map input.
The separate comparison of near-infrared with colour tests whether the band is
redundant.

Replicating a single band across three channels leaves the stem convolution
untouched. Colour, near-infrared, and the luminance control are one network at
one parameter count. The four configurations do not share an annotation:
near-infrared trains on the near-infrared annotation and the other three on the
colour annotation. The input band and that annotation are what vary between the
matched three. The fusion configuration is not matched to them. It runs one
encoder per stream and concatenates the two streams' feature maps stage by
stage. In every model reported here it carries at least \fusionParamFactor{}
times their parameters. The surplus favours fusion. A fusion configuration that
fails to beat the single-band configurations is not failing for want of
capacity.

Every configuration uses the SegFormer semantic-segmentation architecture
\citep{xie2021segformer} with its lightweight all-multi\-layer-perceptron
decoder. The primary encoder is the \backboneSmall{} variant. We repeat the
runs under both protocols with the larger \backboneLarge{} encoder of the same
family to test whether the findings depend on encoder size, and again on
\archSecond{}, which pairs a convolutional backbone with a pyramid-pooling
decoder, to test whether they depend on the architecture family.

Frames are resized to $\inputWidth\times\inputHeight$ pixels and sampled at a
stride of \frameStride{} frames along each trajectory. Training runs on random
square crops of \cropSize{} pixels for \trainEpochs{} epochs, using the AdamW
optimiser at a learning rate of $\learningRate$ decayed to zero on a cosine
schedule, a weight decay of $\weightDecay$, and a batch of \trainBatch{}
frames. We oversample frames containing rare classes using class-balanced
weights, tempered by a square-root law and capped at \samplerCap{}. The
weights are computed once from the union of the colour and near-infrared label
presence and are identical across configurations. 

\subsection{Evaluation protocols and statistical criteria}
The site-blocked protocol trains on a subset of the super-sites and evaluates
on the held-out remainder. No GPS location appears on both sides. The
sequence-level split is the contrasting protocol that admits shared locations.

The leave-one-site-out protocol rotates the held-out super-site across all
\nSuperSites{} folds. Every location serves once as the unseen evaluation
site. The rotation tests whether a site-blocked result depends on which single
site was held out. One driving environment is concentrated in a single
super-site. The fold that holds out that site evaluates a model on an
environment it barely saw in training. One class per rotation must retain an
intersection IoU of at least \controlFloor{} on every fold of a
leave-one-site-out protocol: the farm access road in the unrestricted
rotation, and the tilled paddy surface in the \restrictedEnv{}-restricted
one. A fold on which that class scores below the floor is read as
out-of-domain transfer rather than site generalisation. No leakage-specific
conclusion is drawn from such a fold.

We add a \restrictedEnv{}-restricted leave-one-site-out protocol that confines
both training and evaluation to the driving environment balanced across
super-sites. Holding out a site then tests site generalisation without the
environment shift. To separate site blocking from the environment restriction
itself, we pair this protocol with a matched comparator, the sequence-level
split restricted to the same environment. The matched comparator and the
\restrictedEnv{}-restricted protocol differ only in whether sites are held
apart. Their difference measures site blocking alone.

We also train the colour and near-infrared configurations on one driving
environment at a time under the site-blocked protocol, leaving the evaluation
set at the whole held-out super-site. Each of those models is then scored both
on the environment it trained on and on the one it did not, over the same
frames. The two scores separate a change of environment from a change of site.

Each configuration is scored against \nGroundTruths{} ground truths. The
colour ground truth is the annotation drawn on the colour image, and the
near-infrared ground truth is the annotation drawn on the near-infrared image.
The third is their pixel intersection, in which a pixel belongs to a class
only where the colour and near-infrared annotations agree. The intersection
removes annotation differences specific to one modality. A score on the
intersection cannot be inflated by a configuration matching quirks of its own
annotation. We read every leakage-controlled comparison on the intersection
ground truth for that reason.

Accuracies are reported as a mean over \nSeeds{} random seeds with its
standard deviation, for every configuration and not only for the one a
comparison favours. Per-frame IoU is averaged across seeds within a
configuration before any test. A margin is the mean of per-frame differences.
Subtracting two pooled accuracies does not reproduce it. A frame enters a
comparison by the presence of the class in its annotation rather than by any
model output. The frame set does not depend on the configurations being
compared.

A configuration is judged to beat another on a target class only when the
paired per-frame IoU difference on the intersection ground truth satisfies
three conditions. The paired Wilcoxon signed-rank test rejects at the
\alphaLevel{} level after Holm correction across the \nTargetClasses{} target
classes, the bootstrap 95\% confidence interval on the mean difference excludes
zero, and the difference measured again on a single annotation runs the same
way. That annotation is the near-infrared one where near-infrared is one of the
two configurations and the colour one otherwise. A per-frame difference taken
across the two annotations would carry the difference between the annotations
as well as the one between the configurations. The third condition withholds a
ranking that only the intersection supports. No target class satisfies the
other two and fails it. When any condition fails, the two configurations are
reported as comparable rather than ranked. An effect that also appears on a
control class is reported as suspect, because a control carries no
traversability signal that a spectral band should recover.

A second quantity, the leakage inflation, measures the protocol rather than
the configuration. It is the mean per-frame difference in intersection IoU
between a model trained under a protocol that admits shared locations and a
model trained under a site-disjoint one, paired over the frames both protocols
evaluate. A site-blocked model is scored on the super-site held out from it,
while a model trained on the sequence-level split is scored on that split's
own validation partition. Subtracting the two pooled accuracies would carry
the difference between those two frame sets as well as the leakage. The
paired difference also separates leakage from site difficulty. A held-out
super-site that is harder to segment lowers both models on the same frames,
and only the model that trained on other frames of that super-site gains
from having seen it.

The bootstrap resamples whole driving sequences rather than individual frames.
The autocorrelation that makes a random frame-level split leak also makes a
frame-level interval too narrow, because consecutive frames of one traverse
view the same ground under one exposure setting and are not independent draws.
The sequence is the finest unit our protocols treat as exchangeable and is the
resampling unit throughout. The intervals we report are correspondingly wider
than a frame-level bootstrap would report on the same data. The Wilcoxon test
remains frame-level and is anticonservative at these frame counts. The
interval is the decisive half of the criterion.

Two of the intervals we report end within \ciEndpointFloor{} IoU of zero. We
draw \bootResamples{} resamples, which puts the Monte Carlo error of an
endpoint an order of magnitude below that margin. We mark any criterion whose
resampled mass on the near side of zero sits within \boundarySEs{} Monte Carlo
standard errors of the tail the 95\% interval admits. A marked criterion is
reported as the directional rule returned it and is not read as a result.

The resample count is not the only limit on an endpoint. A cluster bootstrap
produces one value of the mean per multiset of sequences, and an endpoint read
as a percentile of that distribution resolves no more finely than those values
are spaced. On the \suSeqIsec{} sequences behind the
paired contrasts of leakage inflation between configurations, the resampled
mean takes \suSeqIsecMeans{} distinct values.

The evaluation protocol fixes the numeric criteria, the seeds, and the runs to
be made. Every statistic this paper reports is computed under it. The protocol
also requires the study to reproduce an observation from our preliminary
experiments on this corpus, that colour separates the paddy edge better than
near-infrared once the sites are held apart.

The per-class accuracies carry no criterion of their own. They are read under
the directional rule stated here and reported as descriptive. A criterion read
across a rotation is reported on the folds on which the class it names is
evaluable. The paddy edge has no held-out frame at one of the \nSuperSites{}
super-sites. The rotation reports two evaluable folds for that class and no
comparison on the third.

Two analyses depart from the protocol criteria as committed. The frames both
protocols score at the field-heavy super-site are all rice paddy, and the
Holm correction on that fold runs over the two target classes that remain
rather than over all \nTargetClasses{}. The pooled configuration test applies
no correction, although the directional rule includes one.

Beside each criterion that pools a rotation's folds we report a site-level
test, a one-sample $t$ on the \nSuperSites{} fold means. The test is not part
of the criterion. At two degrees of freedom the \alphaLevel{} critical value
is \siteLevelTCrit{} standard errors. A site-level $p$ above \alphaLevel{}
records that the rotation cannot resolve the effect and not that the effect
is absent.

Three measurements taken from the frames, without training anything, test
whether a concentration of leakage in one configuration has a cause other than
the spectral band. The first is exposure. A nearest-centroid classifier given
only the mean and standard deviation of a frame identifies the super-site,
scored by holding out whole driving sequences and read against a majority
baseline, for the near-infrared image and for luma. A configuration whose
brightness alone identifies the site more often has a site signature unrelated
to ground information. The second is supervision. The same classifier is given
a frame's class composition in place of its brightness, once for each of the
two annotations the configurations train on. An annotation that marks the site
more strongly leaves a model more site-specific structure to memorise. The
third is registration. The two modalities come from separate cameras, and an
uncorrected offset between them erodes the intersection ground truth wherever
the two annotations disagree.
Phase correlation on gradient magnitude estimates the offset per frame. The
control for a spurious displacement is the same estimator on pairs whose two
frames come from different driving sequences. The agreement between the two
annotations is then read as scored, at each frame's measured offset, and at
the offset the control returns. The paddy-edge agreement is read again on
the frame sets the comparisons are scored on.

\section{Results}\label{sec:results}

\subsection{No configuration beats colour once the sites are held apart}
The \nEnvironments{} studied environments contribute \num{\nPairedFrames}
frames that carry an annotation on both modalities at the same instant, and
the strided scene-matched index of \num{\nIndexFrames} frames is drawn from
them. The site-blocked protocol holds one super-site out for validation,
\blockedValSharePct{}\% of that index. The nearest pair of frames on opposite
sides of that partition is \SI{\sepCommittedM}{\metre} apart.
Table~\ref{tab:perclass} reports the site-blocked accuracy of each
configuration beside the leakage inflation the sequence-level split admits.

\begin{table}[t]
\centering
\caption{Segmentation accuracy under the site-disjoint protocol, beside the leakage inflation the sequence-level split admits. Accuracies are the intersection-over-union on the intersection ground truth, as a mean over \nSeeds{} seeds with its standard deviation, for the \nModalityArms{} input configurations. The inflation is not the difference of the two blocks. It is the per-frame paired difference between the sequence-level-split and site-blocked models over the frames both protocols scored (Section~\ref{sec:methods}), whereas the two blocks are scored on different frames. Near-infrared carries the largest inflation on 5 of the 6 classes shown. The exception is the field furrow, where fusion is larger. The sequence-level split's own accuracies, the in-plot-tree control that the site-blocked split leaves with nine validation frames, and the non-target standing-rice class are in Table~\ref{tab:perclass-full}.}
\label{tab:perclass}
\footnotesize
\setlength{\tabcolsep}{4pt}
\begin{tabular}{lcccccccc}
\toprule
 & \multicolumn{4}{c}{Site-blocked IoU} & \multicolumn{4}{c}{Leakage inflation} \\
\cmidrule(lr){2-5} \cmidrule(lr){6-9}
Class & Colour & Lum. & NIR & Fusion & Colour & Lum. & NIR & Fusion \\
\midrule
Paddy edge & $0.059_{\pm0.043}$ & $0.011_{\pm0.017}$ & $0.015_{\pm0.016}$ & $0.035_{\pm0.016}$ & $-0.008$ & $+0.027$ & $+0.067$ & $+0.007$ \\
Paddy, untilled & $0.067_{\pm0.016}$ & $0.065_{\pm0.013}$ & $0.100_{\pm0.018}$ & $0.066_{\pm0.005}$ & $+0.051$ & $+0.034$ & $+0.092$ & $+0.060$ \\
Paddy, tilled & $0.507_{\pm0.025}$ & $0.388_{\pm0.037}$ & $0.418_{\pm0.030}$ & $0.553_{\pm0.030}$ & $+0.080$ & $+0.090$ & $+0.104$ & $+0.044$ \\
Field furrow & $0.344_{\pm0.013}$ & $0.253_{\pm0.029}$ & $0.203_{\pm0.014}$ & $0.239_{\pm0.037}$ & $+0.033$ & $+0.060$ & $+0.047$ & $+0.094$ \\
Field levee & $0.538_{\pm0.037}$ & $0.475_{\pm0.022}$ & $0.379_{\pm0.013}$ & $0.539_{\pm0.021}$ & $+0.032$ & $+0.044$ & $+0.129$ & $+0.044$ \\
\midrule
Farm road (control) & $0.832_{\pm0.031}$ & $0.789_{\pm0.011}$ & $0.737_{\pm0.025}$ & $0.847_{\pm0.020}$ & $+0.016$ & $+0.026$ & $+0.033$ & $+0.016$ \\
\bottomrule
\end{tabular}

\vspace{2pt}
\begin{minipage}{\linewidth}\footnotesize Every inflation cell is a descriptive statistic. Which configuration leaks most is established in Section~\ref{sec:concentration}, on contrasts paired between configurations rather than on this column. One cell carries a protocol criterion, near-infrared on the paddy edge, whose sequence-clustered 95\% interval is $[+0.001,\,+0.135]$ over 4 driving sequences. A cluster bootstrap on 4 sequences admits 35 distinct values of the resampled mean, which is how finely those endpoints resolve.\end{minipage}
\end{table}

Under the site-blocked split, near-infrared beats colour on none of the
\nTargetClasses{} target classes, the paddy edge included. Fusing the two
configurations does not recover a consistent gain either. The margins are
paired per-frame means. Fusion exceeds
colour on the tilled paddy surface by \fusionColourBlocked{} (95\% CI
\fusionColourBlockedCI{}), colour exceeds fusion
on the field furrow by \fusionColourFurrow{} (95\% CI \fusionColourFurrowCI{}),
and the remaining three target classes are comparable.

Training on one driving environment at a time does not change the ordering.
Where the training and evaluation environments differ, colour leads
near-infrared on every target cell either configuration scores above zero on
(Table~\ref{tab:transfer}). Two in-domain target cells favour near-infrared,
the field furrow and the field levee under dry-field training, and both
reverse when the runs are scored on each configuration's own annotation rather
than on the intersection.

Near-infrared also exceeds the capacity-matched luminance control on the tilled
paddy surface by \nirLumAfterBlocked{}, but the interval spans zero
(Table~\ref{tab:robustness}b) and the two are comparable. On the untilled
paddy surface the same contrast returns \nirLumBeforeBlocked{} at
\nirLumBeforeBlockedCI{}. It satisfies the directional rule and is
marked. Rotating the held-out super-site across the \nSuperSites{} folds
reverses the sign of this contrast, to \nirLumBeforeSiteLow{} on one of them,
and leaves it indistinguishable from zero at the level of the site ($p =
\nirLumBeforeSiteP$).

The site-blocked result rests on a single held-out super-site. We rotate the
held-out super-site across all \nSuperSites{} folds. A held-out control class,
the farm access road any protocol should segment, remains above its
\controlFloor{} intersection-IoU floor on two folds (\eThreeControlSiteOne{}
and \eThreeControlSiteTwo{}) but is \eThreeControlSiteZero{} on the third. That
fold holds out the super-site containing \fieldShareSiteZeroPct{}\% of all
dry-field frames. The fold is out-of-domain transfer.

Re-clustering the sites cannot rebalance the environment. The minimum
separation between clusters is \SI{\sepCommittedM}{\metre} at the protocol's
\nSuperSites{} super-sites and \SI{\sepPastM}{\metre} one split finer. The
finer clusters are field-free and leave the largest cluster more field-heavy.

Restricting both training and evaluation to rice paddy, the environment
balanced across super-sites, removes the collapse. The tilled paddy surface
remains above its \controlFloor{} intersection-IoU floor on all \nSuperSites{}
folds, the lowest fold mean being \gOneMinFoldMean{}. The
\restrictedEnv{}-restricted rotation supplies \nSuperSites{} held-out sites on
which the tilled-paddy comparisons can be repeated.

Fusion does not retain its advantage over colour, on a post hoc contrast the
criteria for this rotation do not name. The tilled-paddy margin is
\fusionColourSiteTwo{} on the super-site the site-blocked split held out,
\fusionColourSiteOne{} on the second, and \fusionColourSiteZero{} on the
third. Treating each held-out site as one observation gives
\fusionColourSiteMean{} at a site-level $p = \fusionColourSiteP$. The
site-blocked gain changes sign at another super-site.

Near-infrared over the luminance control behaves the same way. Pooling
per-frame differences across the three folds gives \gThreePooled{} (95\% CI
\gThreePooledCI{}, \num{\gThreePooledN} frames), a figure that satisfies the
protocol criterion. The per-fold margins behind that pooled figure are
\gThreeSiteZero{}, \gThreeSiteOne{}, and \gThreeSiteTwo{}, and the same
site-level test gives $p = \gThreeSiteLevelP$. The unrestricted rotation
returns the same pattern. It satisfies the protocol criterion at \eTwoPooled{}
(95\% CI \eTwoPooledCI{}) and gives a site-level $p = \eTwoSiteLevelP$. The
sign reversal between folds withdraws both apparent gains.

The near-infrared margin over the control does not reproduce on either larger
model. Its interval spans zero under both, and under \archSecond{} the margin
has the opposite sign to both SegFormer encoders
(Table~\ref{tab:robustness}b).

Neither departure from the protocol criteria costs a verdict. Forcing the
correction family back to \nTargetClasses{} classes by entering each absent
class at $p = 1$ leaves the outcome of every affected cell unchanged. The
closest of them, colour on the field-heavy fold, exceeds \alphaLevel{} under
both family sizes, and its interval of \eOneRgbSiteZeroCI{} spans zero either
way. The pooled statistics that pass have $p$ values no correction over
\nTargetClasses{} classes could raise past \alphaLevel{}, and the two that fail
do so on the interval rather than on the test. The advantage near-infrared
shows on the paddy edge under the corpus's own split is what remains to be
accounted for.

\subsection{The apparent paddy-edge advantage is site leakage concentrated in
near-infrared}\label{sec:concentration} The sequence-level split places single
locations on both sides of its partition. Its training and validation
sequences share \nOfficialOverlaps{} GPS-footprint overlaps, and every
super-site contributes frames to both partitions. The split evaluates every
model within sites it also trained on. The leakage inflation of
Table~\ref{tab:perclass} is positive on almost every cell, the farm-road
control included. Holding the sites apart lowers more than one class and more
than one configuration. Near-infrared carries the largest inflation on all but
one of the classes shown.

Under the sequence-level split, near-infrared segments the paddy edge at
\nirEdgeOfficial{} intersection IoU against \rgbEdgeOfficial{} for colour, and
under the site-blocked split the same near-infrared accuracy is
\nirEdgeBlocked{} (Tables~\ref{tab:perclass} and~\ref{tab:perclass-full}).
Those two accuracies are scored on different validation frames. The paired
inflation for near-infrared on this class is \nirSeenUnseenIsec{} over the
\num{\nirEdgeIsecFrames} frames both protocols scored. On the paired statistic
near-infrared leads colour by \nirRgbEdgeOfficial{} under the sequence-level
split. Under the site-blocked split the same margin is \nirRgbEdgeBlocked{}
(95\% CI \nirRgbEdgeBlockedCI{}), and its interval reaches zero. No
leakage-controlled outcome placed near-infrared above colour on this class, and
the site-blocked comparison does not establish the colour advantage the
reference states either.

\begin{figure}[t]
\centering
\includegraphics[width=\linewidth]{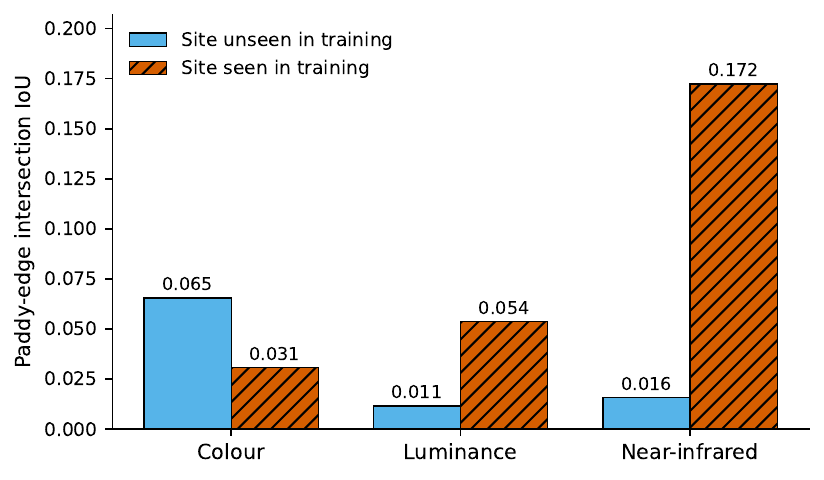}
\caption{Paddy-edge accuracy on a fixed set of frames from one held-out
super-site, scored by two models that differ only in whether they saw the
super-site during training. Bars give the pooled intersection-over-union on the
intersection ground truth for colour, the luminance control, and near-infrared.
Near-infrared gains most from having seen the site and colour least, with the
luminance control between them. The frames are identical and neither model
trained on them. The gap is recognition of the site rather than the site
being harder to segment.}
\label{fig:leakage}
\end{figure}

On the frames of the held-out super-site that neither model trained on, a
model trained on other frames of that super-site scores \nirSeenIsec{} on the
near-infrared paddy edge on the intersection ground truth, against
\nirUnseenIsec{} for a model that never saw the super-site
(Figure~\ref{fig:leakage}). The difference between the two models, averaged
over frames, is \nirSeenUnseenIsec{} for near-infrared, \lumSeenUnseenIsec{}
for the luminance control, and \rgbSeenUnseenIsec{} for colour. The difference
is recognition of the site rather than its difficulty.

The gain from having seen the site is compared across configurations as a
paired quantity. All three are scored on the same frames under the intersection
ground truth. Near-infrared exceeds colour by \suNirRgb{} and exceeds the
capacity-matched luminance control by \suNirLum{}, and the control in turn
exceeds colour by \suLumRgb{}. The interval of every step in that ordering
excludes zero. The leakage is larger for the single-band configurations than
for colour and largest for near-infrared. Those three intervals, \suNirRgbCI{},
\suNirLumCI{} and \suLumRgbCI{}, resample the same \suSeqIsec{} driving
sequences, and those sequences lie inside the single super-site this experiment
holds out. The ordering is resolved within one site and not across sites.

The steps do not hold equally beyond this site and encoder. Near-infrared is
the most inflated of the \nModalityArms{} configurations at the second held-out
super-site. It is also the most inflated configuration under the
\backboneLarge{} encoder, and again under \archSecond{}
(Table~\ref{tab:robustness}a). Both hold the site fixed and vary only the
model. The inflation depends on neither encoder size nor decoder family. The
control's step over colour does not hold. On both larger models colour is no
longer the least inflated configuration and the control is.

The other ground truth does not confirm the top of that ordering. On each
configuration's own annotation the gaps are \lumSeenUnseenOwn{} for the
luminance control, \nirSeenUnseenOwn{} for near-infrared, and
\rgbSeenUnseenOwn{} for colour, placing the control above near-infrared. Those
three gaps cannot all be paired, because near-infrared is scored on its own
annotation and the other two on the colour annotation, and the two annotations
mark the class on different frames. The one contrast that can be paired there
is the control against colour. The two share an annotation. That contrast gives
\suLumRgbOwn{} (95\% CI \suLumRgbOwnCI{}) over \suSeqOwn{} sequences, the same
direction as the intersection contrast but not resolved. Both ground truths
place colour lowest and disagree about which single-band configuration leaks
most, and only the intersection ground truth resolves any step of the
ordering.

The matched comparator for the \restrictedEnv{}-restricted rotation and the
site-disjoint model are scored on the same frames of the held-out super-site.
Neither trained on those frames. Against that comparator, the near-infrared
paddy-edge accuracy inflates by \gTwoNirEdgeSiteZero{} on the field-heavy
super-site, the largest margin of the \nModalityArms{} configurations. The
paddy edge is evaluable on one further super-site of this rotation. There
near-infrared again carries the largest inflation, but no configuration
satisfies the directional rule.

Three mechanisms would explain the concentration without any appeal to the
spectral band, and the frame measurements exclude all three
(Table~\ref{tab:mechanism}). Frame brightness alone identifies the super-site
no better than the majority baseline on near-infrared and better than it on
luma. The annotation supervising near-infrared identifies the super-site no
better than the one supervising the other three. Both are far above that
baseline. The two cameras are not co-registered, at a vertical displacement the
same estimator does not return on frames paired across driving sequences.
Removing each frame's measured offset lifts the paddy-edge agreement on the
corpus-wide sample past what the control's offset reaches, and moves the
large-area classes by little. The erosion is concentrated on thin structure. On
the frames behind the site-blocked comparison and behind the leave-one-site-out
fold the two paddy-edge annotations agree far less than on the corpus-wide
sample, and removing the offset recovers neither. The intersection keeps under
a fifth of the annotated edge pixels on both sets, and the two shares differ by
under a point. On near-infrared's own annotation, which forms no intersection
and carries no cross-camera erosion, the paired inflation on the paddy edge is
\nirSeenUnseenOwn{}, larger than the \nirSeenUnseenIsec{} the same estimator
returns on the intersection ground truth. The paddy-edge advantage under the
sequence-level split is site leakage, and it is concentrated in near-infrared.

\section{Discussion}\label{sec:discussion}

\subsection{The answer for a forward-facing sensor suite}
Near-infrared beat colour on none of the \nTargetClasses{} traversability
target classes once training and evaluation sites were held apart. Both leads
the second camera held at one site, fusion over colour and near-infrared over
the control, reversed at another. The cost of a second camera is not repaid by
a traversability gain that survives cross-site evaluation on this corpus. A
practitioner specifying a forward-facing suite for daytime Korean paddy and
dry-field driving should treat the second camera as unproven for this task.

\subsection{What a leaking split decides, and what a benchmark must do about it}
The sequence-level split, drawn over whole driving sequences, still inflated
near-infrared's paddy-edge accuracy into an apparent advantage over colour.
Separating the recordings is not sufficient on its own. A sensor comparison on
spatially autocorrelated imagery that omits site blocking is an unsafe basis
for a procurement or design decision.

The reordering depends on the leakage being sensor-dependent. A leak that
inflated every configuration alike would preserve their order. Near-infrared
instead memorised the sites more than colour did, and more than the
capacity-matched luminance control did, across both super-sites on which the
paddy edge can be evaluated and across all \nModelConfigs{} model
configurations. The excess over colour is why holding the sites apart removes
near-infrared's lead. The excess over the control places the concentration in
the spectral band and not in the single intensity map near-infrared also
supplies. Frame brightness, the supervising annotation, and the cross-camera
offset produce none of it (Table~\ref{tab:mechanism}). On each configuration's
own annotation the control leaks more than near-infrared. The attribution to
the band rests on the intersection ground truth.

Site blocking removes the leakage but does not by itself license a general
claim. The two leads that reversed under rotation are the evidence. Pooling
per-frame differences over folds does not recover the distinction, because a
pooled figure weights each fold by its frame count. One large site can carry
the pooled pass on its own.

We recommend the four safeguards this study relied on, namely a rotation of
the held-out site tested at the level of the site, a capacity-matched
luminance control, a confidence interval that resamples driving sequences
rather than frames, and a ground truth formed from the intersection of the two
annotations. The sequence-level interval matters because the autocorrelation
that inflates a random-split score narrows a frame-level interval by the same
route. On near-infrared's margin over the luminance control, the frame-level
interval excludes zero. The sequence-level interval on the same data spans it.
A benchmark that argues from spatial autocorrelation must apply the argument
to its own intervals. The intersection removes the leakage that
modality-specific labels carry.

\subsection{Boundaries}
The \nModelConfigs{} model configurations run at one held-out site rule out
encoder capacity and decoder design as the cause of the concentration or of
the margin over the luminance control that spans zero
(Table~\ref{tab:robustness}). Every encoder starts from weights pretrained on
visible-band natural images. Colour is the only configuration whose input
matches that initialisation. One of the two benchmarks whose gains this study
tests also initialises from ImageNet weights and fine-tunes on near-infrared
data \citep{choe2018ranus}. The comparison measures near-infrared under that
initialisation at a budget of \trainEpochs{} epochs. It is not a bound on the
band. The luminance control shares near-infrared's single-map input and its
mismatch to the initialisation. A mismatch that lowers both alike cancels in
the leakage inflation. A configuration whose class signal is harder to read
under this initialisation may instead rely more on whatever identifies the
site. The reliance would concentrate the leakage without the band being its
cause. Encoders pretrained on near-infrared imagery would separate a band
effect from that reliance.

Near-infrared's redundancy with colour is established on one corpus, recorded
in daylight and clear weather in a single province of the Republic of Korea.
The two benchmarks whose gains this study tests also used passive cameras and
describe no night recording \citep{valada2016deep,choe2018ranus}.
The multispectral forest benchmark also reports colour alone above
near-infrared alone, on a different sensor, biome, and label set
\citep{valada2016deep}. That benchmark reports a gain for the band in fusion.
A gain smaller than the site-to-site spread across the \nSuperSites{} held-out
sites is not detectable on this corpus. The balanced site-disjoint rotation
was possible only for rice paddy, because the dry-field environment
concentrates in a single super-site. The larger of the two fusion effects is
on a dry-field class that receives no rotation. The cross-camera offset makes
the thin-class accuracies reported here a floor. It also rules out a per-pixel
vegetation index, because an index presupposes that the two bands are
registered at the pixel where it is read (Table~\ref{tab:mechanism}b).
Replication on further corpora, a balanced site-disjoint benchmark for
dry-field driving, and a corpus with enough distinct sites to resolve an
effect this size are the natural next steps.

\section{Conclusion}\label{sec:conclusion}

The value of a forward-facing near-infrared camera for farm-machinery
traversability depends on whether the band carries ground information that
colour does not. We evaluated \nModalityArms{} input configurations on
site-disjoint splits of the AI Hub \datasetName{} corpus. No configuration
beats colour consistently on any traversability target class. The second
camera's two leads at one site both change sign under rotation of the held-out
site. Near-infrared's advantage under the corpus's own sequence-level split
is recognition of the site. Near-infrared gains more from having seen a site
than colour or a capacity-matched luminance control does. Holding the sites
apart removes the advantage and lowers the other configurations by less. For
daytime paddy and dry-field driving, the second camera is unproven.

\appendix
\counterwithin*{table}{section}

\section{Per-class accuracy under both protocols}\label{app:perclass}

Table~\ref{tab:perclass} reports the site-blocked accuracy of each input
configuration beside the leakage inflation, for the classes compared.
Table~\ref{tab:perclass-full} carries the accuracies themselves under both
protocols, for every annotated class the runs scored. Two classes appear only
there. The in-plot-tree control retains too few validation frames to score
under the site-blocked split, and standing rice is neither a traversability
target nor a control. The sequence-level split's accuracies also appear here
rather than beside the inflation, because the two protocols score different
frames and a reader who subtracted one block from the other would not recover
the paired quantity.

\begin{table}[ht]
\centering
\caption{Per-class segmentation accuracy under the site-blocked protocol and the sequence-level split, for every annotated class the runs scored. Each cell is the intersection-over-union on the intersection ground truth, as a mean over \nSeeds{} seeds with its standard deviation. The site-blocked block repeats Table~\ref{tab:perclass}. The sequence-level-split block is what the leakage inflation there is measured against. In-plot trees are not evaluable under the site-blocked split, which retains nine validation frames of that class. Standing rice is neither a traversability target nor a control and is reported here only for completeness.}
\label{tab:perclass-full}
\footnotesize
\setlength{\tabcolsep}{4pt}
\begin{tabular}{lcccccccc}
\toprule
 & \multicolumn{4}{c}{Site-blocked} & \multicolumn{4}{c}{Sequence-level split} \\
\cmidrule(lr){2-5} \cmidrule(lr){6-9}
Class & Colour & Lum. & NIR & Fusion & Colour & Lum. & NIR & Fusion \\
\midrule
Paddy edge & $0.059_{\pm0.043}$ & $0.011_{\pm0.017}$ & $0.015_{\pm0.016}$ & $0.035_{\pm0.016}$ & $0.009_{\pm0.005}$ & $0.025_{\pm0.014}$ & $0.339_{\pm0.025}$ & $0.055_{\pm0.055}$ \\
Paddy, untilled & $0.067_{\pm0.016}$ & $0.065_{\pm0.013}$ & $0.100_{\pm0.018}$ & $0.066_{\pm0.005}$ & $0.235_{\pm0.047}$ & $0.131_{\pm0.023}$ & $0.239_{\pm0.029}$ & $0.237_{\pm0.033}$ \\
Paddy, tilled & $0.507_{\pm0.025}$ & $0.388_{\pm0.037}$ & $0.418_{\pm0.030}$ & $0.553_{\pm0.030}$ & $0.525_{\pm0.044}$ & $0.491_{\pm0.032}$ & $0.747_{\pm0.040}$ & $0.536_{\pm0.061}$ \\
Field furrow & $0.344_{\pm0.013}$ & $0.253_{\pm0.029}$ & $0.203_{\pm0.014}$ & $0.239_{\pm0.037}$ & $0.366_{\pm0.016}$ & $0.310_{\pm0.022}$ & $0.311_{\pm0.017}$ & $0.332_{\pm0.014}$ \\
Field levee & $0.538_{\pm0.037}$ & $0.475_{\pm0.022}$ & $0.379_{\pm0.013}$ & $0.539_{\pm0.021}$ & $0.661_{\pm0.014}$ & $0.615_{\pm0.022}$ & $0.607_{\pm0.016}$ & $0.658_{\pm0.006}$ \\
\midrule
Farm road (control) & $0.832_{\pm0.031}$ & $0.789_{\pm0.011}$ & $0.737_{\pm0.025}$ & $0.847_{\pm0.020}$ & $0.916_{\pm0.009}$ & $0.848_{\pm0.019}$ & $0.887_{\pm0.018}$ & $0.894_{\pm0.019}$ \\
In-plot trees (control) & \textemdash & \textemdash & \textemdash & \textemdash & $0.329_{\pm0.051}$ & $0.394_{\pm0.035}$ & $0.586_{\pm0.012}$ & $0.388_{\pm0.080}$ \\
Standing rice & $0.642_{\pm0.028}$ & $0.532_{\pm0.043}$ & $0.643_{\pm0.032}$ & $0.551_{\pm0.078}$ & $0.925_{\pm0.008}$ & $0.809_{\pm0.044}$ & $0.820_{\pm0.009}$ & $0.914_{\pm0.012}$ \\
\bottomrule
\end{tabular}
\end{table}

\section{Single-environment training}\label{app:transfer}

Table~\ref{tab:transfer} carries the accuracies of the colour and near-infrared
configurations trained on one driving environment at a time, scored on both
environments of the held-out super-site. Section~\ref{sec:results} states the
ordering they produce. The table carries the accuracies themselves, for every
class with enough held-out frames in an environment to score.

\begin{table}[t]
\centering
\caption{Intersection-ground-truth accuracy of the configurations trained on a single driving environment, evaluated on both environments of the same held-out super-site, as a mean over \nSeeds{} seeds with its standard deviation. The block a row belongs to names the training environment, so the column matching it is in-domain and the other is transfer. Trained on one environment and evaluated on the other, colour leads near-infrared on every target cell either configuration scores above zero on. Near-infrared leads on the intersection ground truth on the field furrow and field levee in domain, and the ordering reverses on the own-label ground truth in every one, so no cell here is read as a configuration advantage. These runs carry no criterion of their own and the accuracies are read as descriptive. A dash marks a class with fewer than \num{50} held-out frames in that environment.}
\label{tab:transfer}
\footnotesize
\setlength{\tabcolsep}{5pt}
\begin{tabular}{lcccc}
\toprule
 & \multicolumn{2}{c}{Evaluated on rice paddy} & \multicolumn{2}{c}{Evaluated on dry field} \\
\cmidrule(lr){2-3} \cmidrule(lr){4-5}
Class & Colour & NIR & Colour & NIR \\
\midrule
\multicolumn{5}{l}{\emph{Trained on rice paddy alone}} \\
Paddy edge & $0.207_{\pm0.096}$ & $0.079_{\pm0.022}$ & \textemdash & \textemdash \\
Paddy, untilled & $0.000_{\pm0.000}$ & $0.000_{\pm0.000}$ & \textemdash & \textemdash \\
Paddy, tilled & $0.685_{\pm0.031}$ & $0.585_{\pm0.022}$ & $0.870_{\pm0.031}$ & $0.649_{\pm0.055}$ \\
Field furrow & $0.115_{\pm0.039}$ & $0.000_{\pm0.000}$ & $0.024_{\pm0.016}$ & $0.000_{\pm0.000}$ \\
Field levee & $0.359_{\pm0.031}$ & $0.000_{\pm0.000}$ & $0.162_{\pm0.043}$ & $0.000_{\pm0.000}$ \\
Farm road (control) & $0.808_{\pm0.020}$ & $0.715_{\pm0.044}$ & $0.907_{\pm0.008}$ & $0.838_{\pm0.039}$ \\
Standing rice & $0.709_{\pm0.021}$ & $0.585_{\pm0.045}$ & \textemdash & \textemdash \\
\midrule
\multicolumn{5}{l}{\emph{Trained on dry field alone}} \\
Paddy edge & $0.000_{\pm0.000}$ & $0.000_{\pm0.000}$ & \textemdash & \textemdash \\
Paddy, untilled & $0.080_{\pm0.005}$ & $0.048_{\pm0.018}$ & \textemdash & \textemdash \\
Paddy, tilled & $0.104_{\pm0.020}$ & $0.036_{\pm0.030}$ & $0.584_{\pm0.059}$ & $0.353_{\pm0.048}$ \\
Field furrow & $0.175_{\pm0.028}$ & $0.059_{\pm0.041}$ & $0.348_{\pm0.013}$ & $0.399_{\pm0.019}$ \\
Field levee & $0.231_{\pm0.050}$ & $0.109_{\pm0.014}$ & $0.704_{\pm0.007}$ & $0.711_{\pm0.019}$ \\
Farm road (control) & $0.757_{\pm0.024}$ & $0.673_{\pm0.031}$ & $0.891_{\pm0.016}$ & $0.895_{\pm0.005}$ \\
Standing rice & $0.000_{\pm0.000}$ & $0.000_{\pm0.000}$ & \textemdash & \textemdash \\
\bottomrule
\end{tabular}
\end{table}

\section{Encoder size and architecture family}\label{app:robustness}

Table~\ref{tab:robustness} carries both cross-model contrasts of
Section~\ref{sec:results} on every input configuration and every model, namely
the paddy-edge leakage inflation and the near-infrared accuracy margin over the
capacity-matched luminance control on the tilled paddy surface. The
near-infrared cell of each is where the protocol criteria are declared. The
table carries every cell.

\begin{table}[ht]
\centering
\caption{Encoder size and architecture family. Both panels are read on the frames the site-blocked split holds out, so the model varies and the held-out site does not. What they test is whether an ordering depends on the encoder or the decoder, not whether it holds at another site. (a) The paddy-edge leakage inflation, the per-frame paired difference between the sequence-level-split and site-blocked models over the \num{322} frames both protocols scored, for the \nModalityArms{} input configurations. Near-infrared carries the largest inflation on all \nModelConfigs{} models. The interval is on the near-infrared cell, the one cell of the panel a protocol criterion is declared on. The others are descriptive. (b) The near-infrared accuracy margin over the capacity-matched luminance control on the tilled paddy surface under the site-blocked protocol. The margin spans zero on all \nModelConfigs{} models, and has the opposite sign on \archSecond{}. Intervals are sequence-clustered 95\% bootstrap intervals.}
\label{tab:robustness}
\footnotesize
\setlength{\tabcolsep}{4pt}
\emph{(a) Paddy-edge leakage inflation}\par\vspace{2pt}
\begin{tabular}{lccccc}
\toprule
Model & Colour & Lum. & NIR & Fusion & 95\% CI, NIR \\
\midrule
\backboneSmall{} (primary) & $-0.008$ & $+0.027$ & $+0.067$ & $+0.007$ & $[+0.001,\,+0.135]$ \\
\backboneLarge{} & $+0.042$ & $+0.041$ & $+0.100$ & $+0.046$ & $[+0.034,\,+0.175]$ \\
\archSecond{} & $+0.038$ & $+0.029$ & $+0.069$ & $+0.032$ & $[+0.001,\,+0.157]$ \\
\bottomrule
\end{tabular}
\par\vspace{8pt}
\emph{(b) Near-infrared over the luminance control on the tilled paddy surface}\par\vspace{2pt}
\begin{tabular}{lcc}
\toprule
Model & NIR $-$ Lum. & 95\% CI \\
\midrule
\backboneSmall{} (primary) & $+0.030$ & $[-0.081,\,+0.104]$ \\
\backboneLarge{} & $+0.019$ & $[-0.104,\,+0.106]$ \\
\archSecond{} & $-0.071$ & $[-0.168,\,+0.014]$ \\
\bottomrule
\end{tabular}
\end{table}

\section{What does not explain the concentration}\label{app:mechanism}

Table~\ref{tab:mechanism} carries the three mechanism measurements of
Section~\ref{sec:concentration}: whether frame brightness alone identifies the
super-site, whether the annotation a configuration is supervised by identifies
it, and how far the cross-camera offset erodes the paddy-edge intersection on
the frame sets the comparisons are scored on. None is a result the paper
claims. Each closes an explanation of the concentration that would not need the
spectral band, and the section states what each returned.

\begin{table}[ht]
\centering
\caption{The three mechanisms that would explain the concentration of leakage in near-infrared without any appeal to the spectral band, all measured from the frames without training anything. (a) Super-site identified by one nearest-centroid classifier, scored by holding out whole driving sequences over the \num{15790} frames of \num{94} sequences, from the mean and standard deviation of the frame and from the class composition of the annotation the configuration is supervised by. The luminance control's row reads the colour annotation, which supervises colour and fusion too. Near-infrared does not exceed the majority baseline on brightness and luma does, so near-infrared brightness carries no site signature colour brightness lacks. Both annotations exceed that baseline by a wide margin and differ from each other by \annotSiteIdDelta{} (95\% CI \annotSiteIdDeltaCI{}), so the annotation supervising near-infrared marks the site no more strongly than the one supervising the other three. (b) Median agreement between the two paddy-edge annotations, as scored and after each frame's measured camera offset is removed, against the same agreement at the control's offset, and with the share of annotated edge pixels the intersection keeps. The two cameras are a median \num{\camShiftPx}~px apart vertically at the working resolution, against \num{\camShiftControlPx}~px for the same estimator on frames paired across driving sequences, which is what it reports from no alignment at all. The control's offset is read on the corpus sample, where alignment is claimed to recover something. Alignment lifts the corpus sample and leaves both comparison sets below even its unaligned agreement. The corpus sample is stratified over the whole corpus and is not what any comparison is scored on. The other two rows are the frame sets a comparison is scored on, and they are eroded to within a point of each other, which is what lets an ordering that agrees across both be read as more than an annotation artifact.}
\label{tab:mechanism}
\footnotesize
\setlength{\tabcolsep}{4pt}
\emph{(a) Exposure and supervision}\par\vspace{2pt}
\begin{tabular}{lcc}
\toprule
& \multicolumn{2}{c}{Site-identification accuracy} \\
\cmidrule(lr){2-3}
Configuration & Frame brightness & Supervising annotation \\
\midrule
Near-infrared & $0.440$ & $0.646$ \\
Luminance control & $0.534$ & $0.640$ \\
Majority baseline & $0.472$ & $0.472$ \\
\bottomrule
\end{tabular}
\par\vspace{8pt}
\emph{(b) Registration}\par\vspace{2pt}
\begin{tabular}{lccccc}
\toprule
Frame set & Frames & Agreement & Aligned & At control & Intersection keeps \\
\midrule
Corpus sample & $232$ & $0.470$ & $0.780$ & $0.475$ & --- \\
Site-blocked comparison & $322$ & $0.082$ & $0.004$ & --- & $19.5$\% \\
Leave-one-site-out fold & $256$ & $0.136$ & $0.169$ & --- & $18.8$\% \\
\bottomrule
\end{tabular}
\end{table}

\bibliographystyle{elsarticle-num-names}
\bibliography{references}

\end{document}